\documentclass[runningheads]{llncs}

\usepackage{eccv}

\usepackage{eccvabbrv}

\usepackage{graphicx}
\usepackage{booktabs}
\usepackage{tikz}
\usepackage{pgfplots}
\pgfplotsset{compat=1.18}
\usepackage{caption}
\usepackage{subcaption} 
\usepackage{wrapfig}
\usepackage{tabularx}
\usepackage{makecell}

\usepackage[accsupp]{axessibility}  

\usepackage{hyperref}

\usepackage{orcidlink}

\usepackage{multirow}
\begin{document}

\title{Understanding Geometric Representations in Self-Supervised Vision Transformers via Subspace Intervention}

\titlerunning{Understanding ViTs via Subspace Intervention}

\author{
Weichen Zhou \inst{1}  \and
Yawen Zou    \inst{1}  \and
Chunzhi Gu   \inst{2}  \and
Ran Dong     \inst{3}  \and
Haoran Xie   \inst{4}  \and
Chao Zhang   \inst{1}  \thanks{Corresponding author.}
}

\authorrunning{W. Zhou et al.}

\institute{
University of Toyama, Toyama, Japan
\and
University of Fukui, Fukui, Japan
\and
Chukyo University, Aichi, Japan
\and
Japan Advanced Institute of Science and Technology (JAIST), Ishikawa, Japan
}

\maketitle


\begin{abstract}
We introduce a controlled subspace intervention framework to investigate how self-supervised Vision Transformers (ViTs) encode dense geometric information. While linear probing is widely used to assess geometric representations, it treats features as a black box, failing to disentangle the underlying topology. To address this issue, we decompose the weights of converged linear probes to isolate the low-rank subspaces containing explicit geometric signals using Singular Value Decomposition (SVD). Our perspective yields three key insights: (1) Pre-training objectives determine how features are encoded. DINOv2 aligns spatial features for efficient linear extraction, while Masked Autoencoders (MAE) tend to disperse these signals, requiring a broader spatial context. (2) Explicit geometric representations are highly compressible, suggesting dense predictive heads could potentially be constrained to low-rank subspaces with minimal performance loss. (3) The layer-wise task affinity suggests that geometric precision peaks at intermediate layers before yielding to semantic abstraction in the final layers. By connecting internal encoding mechanics with downstream performance, these findings provide a basis for effective feature selection and lightweight decoder design. The source code is available at \url{https://github.com/Zhou-Weichen/Geosubprobe}.

\end{abstract}


\section{Introduction}
\label{sec:intro}

Modern dense prediction frameworks increasingly rely on self-supervised Vision Transformers (ViTs) as their foundational backbones~\cite{Yang2024DepthAU,Yang2024DepthAV, Jevtic2025FeedForwardSF,Wang2025VGGTVG}. Beyond the expected gains from task-specific training, these models exhibit an inherent understanding of spatial geometric information directly within their pre-trained representation space~\cite{Oquab2023DINOv2LR,Amir2021DeepVF,Mao2024StealingSD}. While their empirical effectiveness is established, the encoding format of this emergent geometric information remains unclear. It is uncertain whether the network disperses geometric information across its entire high-dimensional feature space or aligns it into specific, linearly accessible coordinates. Drawing from insights that latent space topology is heavily influenced by pre-training objectives~\cite{Wang2020UnderstandingCR, Xie2022RevealingTD, Park2023WhatDS}, we hypothesize that optimization constraints determine how geometric primitives are routed and formatted within latent subspaces. Understanding how distinct paradigms, such as self-distillation~\cite{Oquab2023DINOv2LR}, masked image modeling~\cite{He2021MaskedAA}, and hybrid approaches~\cite{Zhou2021iBOTIB}, shape these internal geometric representations is therefore critical for adapting these models to downstream tasks.

To characterize this inherent geometric understanding, recent studies reveal that self-supervised ViTs encode accurate single-view geometry despite lacking explicit 3D supervision~\cite{ElBanani2024ProbingT3,Zhan2023AGP,Man2024Lexicon3DPV,Chen2024ProbingTM}. This geometric awareness is a functional necessity for resolving complex semantic ambiguities~\cite{Zhang2023TellingLF}. However, these models often struggle with multi-view consistency and complex spatial relationships~\cite{Man2024Lexicon3DPV, Chen2024ProbingTM}. Such inconsistencies suggest that simply detecting the presence of geometric information is insufficient; it is crucial to understand how it is topologically encoded. Currently, representations are assessed primarily through the final prediction accuracy of black-box downstream probes~\cite{ElBanani2024ProbingT3, Chen2024ProbingTM, Alain2016UnderstandingIL, Belinkov2021ProbingCP}. This paradigm conflates the absence of information with the inability to decode it~\cite{Hewitt2019DesigningAI,Pimentel2020InformationTheoreticPF}. When a linear probe performs poorly, it remains ambiguous whether the geometric information is absent, entangled within non-linear manifolds, or scattered across disjoint spatial patches. Resolving this diagnostic ambiguity requires directly examining the underlying feature encoding mechanics.

To address these uncertainties, we introduce a controlled subspace intervention framework. Since a linear probe extracts information by projecting features onto a learned weight matrix, its predictive capacity is bottlenecked by the continuous subspace spanned by these weights. Motivated by this property, we apply SVD to the converged weights of linear probes to explicitly isolate task-aligned geometric directions. This allows us to evaluate the encoding format where geometric signals recoverable from a low-dimensional subspace indicate highly aligned spatial features. In contrast, signals requiring high-dimensional non-linear aggregation across spatial tokens should be entangled.

Through this analytical framework, our investigation reveals distinct feature encoding formats driven by pre-training objectives. Self-distillation (\eg, DINOv2) strongly aligns explicit geometry into highly compressible, low-rank coordinate systems. In contrast, generative masked reconstruction model (\eg, MAE) disperses geometric signals across broader dimensions to satisfy pixel-level reconstruction constraints. Despite these differences, we identify a high compressibility of explicit geometric representations across all evaluated paradigms. Furthermore, layer-wise evaluation uncovers a task affinity among surface normal estimation, depth estimation, and semantic segmentation in the deeper representation space.

Our main contributions are summarized as follows:
\begin{itemize}
    \item We introduce a controlled subspace intervention framework. By applying SVD to converged linear probe weights, we isolate task-aligned directions to quantify the linear compressibility of geometric representations, moving beyond black-box output metrics.
    
    \item Through our framework, we characterize differences in the accessibility of geometric information across the evaluated self-supervised paradigms. DINOv2 exhibits more linearly accessible geometry than MAE, while much of the linearly decodable geometric information is concentrated in low-rank, task-aligned subspaces.
    
    \item We reveal that geometric features peak in DINOv2's middle layers, then weaken as semantic performance improves.
    
\end{itemize}


\begin{figure}[tb]
\centering
\includegraphics[width=\textwidth]{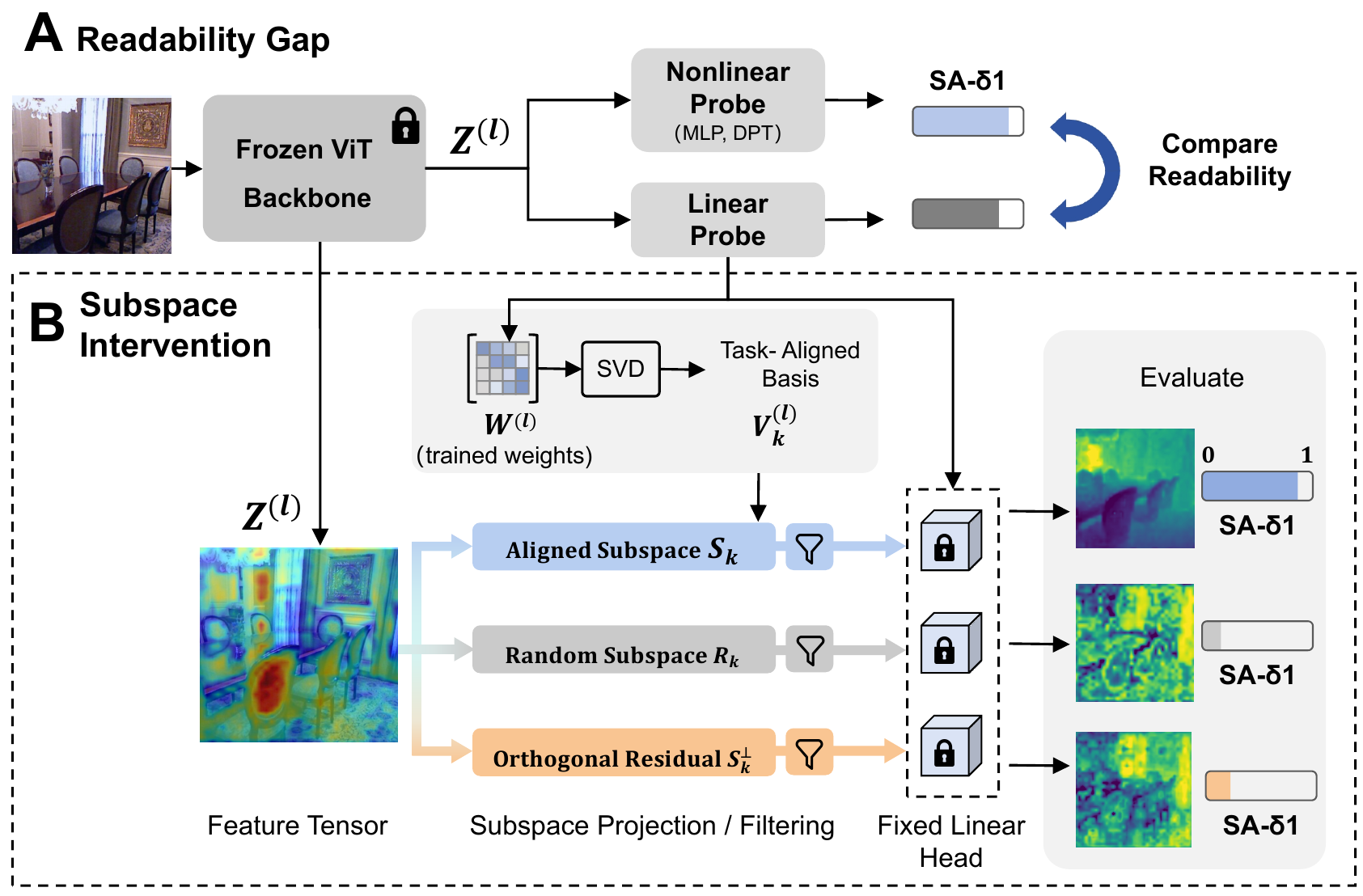} 
\caption{\textbf{Overview of the controlled subspace intervention analysis framework.} (A) We evaluate frozen backbone features $\mathbf{Z}$ using three-tier probes (Linear, MLP, and DPT) to decouple local non-linear entanglement from global spatial fragmentation, thereby obtaining the readability gap of geometric features. (B) Without additional training, we perform SVD on the converged linear weights $\mathbf{W}$ to extract a task-aligned basis $\mathbf{V}_k$. The feature tensor is then projected onto the aligned subspace ($\mathcal{S}_k$), a random subspace ($\mathcal{R}_k$), and the orthogonal residual ($\mathcal{S}_k^\perp$). The projected features are evaluated through the fixed linear head to isolate the geometric signal.}
\label{fig:framework}
\end{figure}

\section{The Diagnostic Framework}
\label{sec:framework}
Traditional evaluation paradigms assess representations through the black-box probing of downstream outputs, offering limited insights into the underlying feature structure. We introduce a post-hoc analytical framework that transitions from only evaluating task accuracy to inspecting the internal manifold topology. As illustrated in \cref{fig:framework}, our approach consists of two primary stages: establishing a readability gap (Panel A) and performing subspace interventions (Panel B).

\subsection{Readability Gap}
\label{sec:readability_gap}
Our first objective is to determine whether geometric information is genuinely absent or locked within complex non-linear structures. Rather than using a single decoder, we design a three-tier probing hierarchy. By deliberately restricting the capacity of these probe heads, we decouple local non-linear folding from the necessity of global spatial context. The resulting performance discrepancies across these tiers define the readability gap, providing a baseline of topological entanglement before we delve into its subspace.

\subsubsection{\textit{Tier1}: Linear Probing (Explicit Geometry).} Given the patch-level representation $\mathbf{Z}^{(l)} \in \mathbb{R}^{N \times D}$ from layer $l$ of a frozen backbone, where $N$ denotes the number of spatial tokens and $D$ represents the feature dimension per token, we apply a linear head parameterized by $\mathbf{W}^{(l)} \in \mathbb{R}^{C \times D}$, where $C$ is the target prediction space dimensionality (\eg, $C=256$ depth bins~\cite{Bhat2020AdaBinsDE}). The prediction is computed as $\hat{\mathbf{Y}} = \mathbf{Z}^{(l)} \mathbf{W}^{(l)\top}$. This naturally extends to a concatenated multi-scale representation mapped by $\mathbf{W}_{\mathrm{global}} \in \mathbb{R}^{C \times4D}$. We treat this linear probe as a strict measure of accessibility, testing whether geometric cues are explicitly rectified into a linearly readable format without requiring cross-patch communication. 

\subsubsection{\textit{Tier2}: MLP and \textit{Tier3}: DPT Nonlinear Controls (Entanglement vs. Fragmentation).} When a linear probe fails, it remains unclear whether the geometry is entirely absent, non-linearly folded within individual patches, or scattered across the spatial manifold. To isolate this, we introduce a token-wise Multi-Layer Perceptron (MLP) and a DPT-style decoder~\cite{Ranftl2021VisionTF}. The MLP operates point-wise on individual spatial tokens but incorporates non-linear activations, so the performance gap between the Linear and MLP probes quantifies the degree of local non-linear entanglement. The performance gap between the DPT and MLP probes isolates the degree of spatial dispersion, revealing whether the backbone distributes geometric primitives across multiple spatial tokens that necessitate a global receptive field for inference.  Together, this probing hierarchy establishes a comprehensive readability description.

\subsection{Subspace Intervention}
\label{sec:subspace}

While the readability gap quantifies the degree of entanglement, it does not reveal the directional distribution of information. To peer inside the linear probe's mechanism, we introduce a subspace intervention post-hoc analysis framework.

\subsubsection{SVD on Probe Weights.} We posit that the converged linear weight matrix $\mathbf{W}^{(l)}$ encodes the task-aligned geometric directions discovered by the probe. Because the mapping projects to a low-dimensional target space ($C \ll D$), the rank of the learned linear mapping is upper-bounded by $C$. We perform SVD directly on the probe weights without requiring additional training: 
\begin{equation}
\mathbf{W}^{(l)} = \mathbf{U}^{(l)} \boldsymbol{\Sigma}^{(l)} \mathbf{V}^{(l)\top}
\end{equation}
where $\mathbf{U}^{(l)} \in \mathbb{R}^{C \times C}$ contains the left singular vectors, $\boldsymbol{\Sigma}^{(l)} \in \mathbb{R}^{C \times C}$ is a diagonal matrix containing the singular values, and $\mathbf{V}^{(l)} \in \mathbb{R}^{D \times C}$ contains the right singular vectors. The columns of $\mathbf{V}^{(l)}$ form an orthonormal basis of the task-relevant representation space. By extension, the same decomposition is applied to the global concatenated weight matrix $\mathbf{W}_{\mathrm{global}}$, yielding a corresponding low-rank basis spanning the multi-scale feature dimensions.

\subsubsection{Aligned Subspace Identification.}
We define aligned subspace of rank $k$ ($k \le C$) as the span of the top-$k$ principal directions:
\begin{equation}
\mathcal{S}^{(l)}_k = \mathrm{span}\{\mathbf{v}_1, \dots, \mathbf{v}_k\}.
\end{equation}
To assess the geometric density of this subspace, we constrain the backbone representation via orthogonal projection:
\begin{equation}
\tilde{\mathbf{Z}}^{(l)}_k = \mathbf{Z}^{(l)} \mathbf{V}^{(l)}_k \mathbf{V}^{(l)\top}_k,
\end{equation}
Crucially, the projected features are evaluated directly using the original, frozen linear head. Because both the backbone and the probe remain locked, any performance variation is exclusively attributable to the geometric capacity of the selected $k$-dimensional subspace. To ensure that the extracted basis reflects the intrinsic manifold rather than optimization artifacts, we evaluated the similarity of the extracted subspaces across multiple random probe initializations (see Section~\ref{sec:robustness}).

\subsection{Control Subspaces: Isolating the Signal}
\label{sec:ablation}

To ensure that the explicit geometric signal is directionally concentrated along the principal components, rather than an artifact of retaining $k$ degrees of freedom, we introduce two control subspaces for strict ablation under the frozen probe:

First, the random subspace $\mathcal{R}_k$ projects features onto a span of $k$ randomly sampled orthonormal directions in $\mathbb{R}^D$ or $\mathbb{R}^{4D}$. This disrupts the task-aligned structure while preserving the exact dimensionality, thereby serving as a null baseline. Second, to isolate the orthogonal complement containing the rejected tail dimensions, we compute the residual representation $\mathbf{Z}^{(l)}_{\mathrm{res}}$ by subtracting the task-aligned projection from the original features:
\begin{equation}
\mathbf{Z}^{(l)}_{\mathrm{res}} = \mathbf{Z}^{(l)} - \mathbf{Z}^{(l)} \mathbf{V}^{(l)}_k \mathbf{V}^{(l)\top}_k.
\end{equation}
Passing these residual features through the frozen head allows us to validate the sufficiency of the low-rank subspace and assess the information loss in its orthogonal complement.

Evaluating these residual representations through the same frozen linear probe serves as a diagnostic for whether the task-aligned geometric signal is concentrated in the top-$k$ subspace. A sharp decrease in performance on the residual space indicates that the explicitly decodable geometric features exploited by the probe are largely confined to the top-$k$ components, rather than being broadly dispersed across the remaining dimensions. This experimental design determines whether the network compresses geometric signals into a compact manifold or relies on a more dispersed distribution.

\section{Experimental Design and Objectives}
\label{sec:experiments}

\subsection{Datasets and Implementations}
\label{sec:exp_setup}
We evaluate three representative self-supervised ViT paradigms: self-distillation (DINOv2~\cite{Oquab2023DINOv2LR}), masked image modeling (MAE~\cite{He2021MaskedAA}), and a hybrid approach (iBOT~\cite{Zhou2021iBOTIB}). For our primary analysis, we adopt the ViT-Large variant for all architectures and maintain frozen backbone weights.

Geometric representations are primarily assessed via monocular depth estimation on the standard NYU Depth V2 dataset~\cite{Silberman2012IndoorSA}. Performance is measured by scale-aware (SA) threshold accuracy ($\delta < 1.25$), RMSE, and Absolute Relative error (AbsRel)~\cite{Eigen2014DepthMP}. To investigate layer-wise task affinities within a controlled domain, we perform parallel probing for 40-class semantic segmentation~\cite{Cao2021ShapeConvSC} and surface normal estimation, adopting the mean Intersection over Union (mIoU) and angular accuracy~\cite{bae2021estimating}($d_1 < 11.25^\circ$) as the respective evaluation metrics.

Since one of the main contributions of this work is the analytical framework itself, which utilizes converged probe weights to inversely explore the internal feature topology, we focus on depth estimation to construct a cohesive and deep narrative. However, to examine the consistency of findings across tasks, domains, and probe configurations, we provide extensive parallel evaluations in the Supplementary Material. These include full surface normal trajectories, extended cross-domain benchmarks (\eg, depth on KITTI~\cite{geiger2012we,geiger2013vision}, and depth/normal estimation on NAVI~\cite{jampani2023navi}), and validation on ViT-Base variants.

\subsection{The Three-tier Probing Hierarchy}
\label{sec:three_tier}
As formalized in \cref{sec:framework}, we implement a three-tiered probing hierarchy to systematically distinguish the presence of geometric information from its structural accessibility. The baseline \textbf{Linear Probe} employs a $1 \times 1$ convolution to map frozen patch tokens directly to the target space, quantifying explicit, isolated geometric cues. To isolate local non-linear entanglement, the intermediate \textbf{MLP Probe} introduces point-wise non-linear activations (\eg, a hidden layer with GELU) while strictly maintaining the exact same $1 \times 1$ token-wise receptive field. Finally, the \textbf{DPT Decoder}~\cite{Ranftl2021VisionTF} establishes the absolute latent geometric upper bound by incorporating multi-scale feature aggregation and global spatial receptive fields.

\subsection{Analytical Roadmap}
\label{sec:roadmap}
Our experimental analysis follows a deductive trajectory, including each stage and its corresponding objective.

(1) \cref{sec:topology}: We assess the readability gap by applying different decoders (Linear, MLP, and DPT) to concatenated multi-layer features to quantify feature alignment versus dispersion.

(2) \cref{sec:global_subspace}: Building upon the decoder baselines in (1), we apply SVD to the global probe weights to evaluate the compressibility of geometric representations and quantify their cross-layer energy routing.

(3) \cref{sec:phase_transition}: Motivated by energy imbalances, we investigate rank sensitivity at individual network depths through single-layer subspace interventions.

(4) \cref{sec:semantic_tradeoff}: We contrast geometric and semantic probing to examine whether terminal performance shifts originate from capacity loss or a transition in task-specific affinities.

(5) \cref{sec:robustness}: We verify the stability of the extracted subspaces across random probe initializations.

Full implementation details, including hyperparameters, optimizer, and exact architectural specifications, are provided in the Supplementary Material.


\section{Results and Analysis}
\label{sec:results}

\subsection{Feature Encoding : Alignment vs. Dispersion}
\label{sec:topology}

\begin{table}[t]
\centering
\caption{\textbf{Linear Readability Gap Analysis.} Comparing the geometric performance (SA-$\delta_1$) across our three-tier probing. The decomposition of gaps isolates point-wise non-linear folding  (Local Entanglement) from reliance on global context (Spatial Fragmentation), quantifying the degree of spatial feature alignment versus dispersion.}
\label{tab:baseline_gap}
\small 
\begin{tabularx}{\linewidth}{l *{3}{>{\centering\arraybackslash}X} *{2}{>{\centering\arraybackslash}X}}
\toprule
 & \multicolumn{3}{c}{\textbf{Probe Architecture (SA-$\delta_1 \uparrow$)}} & \multicolumn{2}{c}{\textbf{Accessibility Gaps}} \\
\cmidrule(lr){2-4} \cmidrule(lr){5-6}
\textbf{Backbone} & \textbf{Linear} & \textbf{1$\times$1 MLP} & \textbf{DPT} & \makecell{\textbf{Local}\\\textbf{Entang.}} & \makecell{\textbf{Spatial}\\\textbf{Frag.}} \\
\midrule
\textbf{DINOv2-L} & 0.9157 & 0.9325 & 0.9483 & +0.0168 & +0.0158 \\
\addlinespace
\textbf{iBOT-L}   & 0.8198 & 0.8376 & 0.8524 & +0.0178 & +0.0148 \\
\addlinespace
\textbf{MAE-L}    & 0.6033 & 0.6390 & 0.7022 & +0.0357 & +0.0632 \\
\bottomrule
\end{tabularx}
\end{table}

As shown in \cref{tab:baseline_gap}, evaluating the three pre-training paradigms via our tiered probing reveals distinct feature encoding formats, distinguishing geometric representation presence from structural accessibility for downstream decoders.

DINOv2 exhibits strong spatial feature alignment. A linear probe alone recovers the vast majority of geometric signals (SA-$\delta_1$ of 0.9157). Point-wise MLP probing improves this slightly to 0.9325, while the global DPT decoder yields 0.9483.  This indicates that self-distillation inherently organizes geometric information into a linearly accessible format, minimizing the need for complex downstream decoders. In contrast, MAE demonstrates significant spatial dispersion. Linear performance drops to 0.6033, and MLP probing only raises accuracy to 0.6390. However, the DPT decoder's global receptive field substantially boosts performance to 0.7022.  This jump (+0.0989 over Linear) indicates that masked reconstruction distributes geometric primitives across patches, necessitating global spatial aggregation for effective extraction. The hybrid iBOT model represents an intermediate state. Its minimal performance gains from the 0.8198 linear baseline to the 0.8524 DPT indicate low non-linear entanglement and fragmentation, suggesting that hybrid objectives balance linear accessibility with dense localized representations.

\begin{figure}[t]
\centering
\begin{subfigure}{0.48\linewidth}
\centering
\begin{tikzpicture}
\begin{axis}[
    width=\linewidth,
    height=5.5cm,
    xlabel={Subspace Rank ($k$)},
    ylabel={SA-$\delta_1$ Accuracy},
    xmin=0, xmax=135,
    ymin=0, ymax=1,
    xtick={4,32,64,128}, 
    extra x ticks={16},
    extra x tick labels={16},
    ymajorgrids=true,
    grid style=dashed,
    tick label style={font=\small},
    label style={font=\small},
    legend style={at={(0.97,0.22)}, anchor=south east, font=\tiny, row sep=-2pt},
    mark size=1.5pt,
    axis on top
]
\fill[gray!20] (axis cs:0,0) rectangle (axis cs:135,0.17);
\node[anchor=north, gray!80!black] at (axis cs: 67, 0.17) {
    \scalebox{0.9}{
        $\longleftarrow$ Residual / Random  $\longrightarrow$
    }
};
\addplot[color=blue, mark=square*, thick] coordinates {(4,0.0505)(8,0.3430)(16,0.6973)(32,0.6959)(64,0.7775)(128,0.8661)};
\addlegendentry{DINOv2}
\addplot[color=purple, mark=triangle*, thick] coordinates {(4,0.2655)(8,0.2728)(16,0.6524)(32,0.7511)(64,0.7790)(128,0.7980)};
\addlegendentry{iBOT}
\addplot[color=orange, mark=*, thick] coordinates {(4,0.3092)(8,0.4571)(16,0.5423)(32,0.5908)(64,0.6041)(128,0.6036)};
\addlegendentry{MAE}

\end{axis}
\end{tikzpicture}
\caption{}
\label{fig:subspace_sig_noise}
\end{subfigure}
\hfill
\begin{subfigure}{0.48\linewidth}
\centering
\begin{tikzpicture}
\begin{axis}[
    width=\linewidth,
    height=5.5cm,
    xlabel={Subspace Rank ($k$)},
    ylabel={Recovery Ratio},
    xmin=0, xmax=135,
    ymin=0, ymax=1.1,
    xtick={4,32,64,128},
    ytick={0, 0.2, 0.4, 0.6, 0.8, 1.0},
    extra x ticks={16},
    ymajorgrids=true,
    grid style=dashed,
    tick label style={font=\small},
    label style={font=\small},
    legend style={at={(0.97,0.08)}, anchor=south east, font=\tiny, row sep=-2pt},
    mark size=1.5pt
]
\addplot[color=blue, mark=square*, thick] coordinates {(4,0.055)(8,0.375)(16,0.761)(32,0.760)(64,0.849)(128,0.946)};
\addlegendentry{DINOv2}
\addplot[color=purple, mark=triangle*, thick] coordinates {(4,0.324)(8,0.333)(16,0.796)(32,0.916)(64,0.950)(128,0.973)};
\addlegendentry{iBOT}
\addplot[color=orange, mark=*, thick] coordinates {(4,0.512)(8,0.758)(16,0.899)(32,0.979)(64,1.001)(128,1.000)};
\addlegendentry{MAE}
\end{axis}
\end{tikzpicture}
\caption{}
\label{fig:subspace_efficiency}
\end{subfigure}
\caption{\textbf{(a) Absolute Recovery:} Solid colored lines depict the performance of the task-aligned subspace ($\mathbf{V}_k \mathbf{V}_k^\top$). The gray shaded area denotes the noise floor ($\text{SA-}\delta_1 < 0.18$), accounting for both the residual subspace and the random orthogonal baselines under the frozen probe. This gap indicates substantial representational redundancy, suggesting that explicitly decodable geometric information can be compressed into a low-rank subspace without significant loss.
\textbf{(b) Recovery Efficiency:} Normalized against each model's full-rank linear baseline, MAE (Orange) exhibits the fastest relative saturation, recovering $>98\%$ of its linear potential by $k=32$. iBOT (Purple) tracks closely with MAE, while DINOv2 (Blue) shows slower convergence. This indicates that DINOv2 linearly encodes richer, fine-grained geometric details that require slightly more dimensions to fully resolve.}
\label{fig:subspace_pair}
\end{figure}
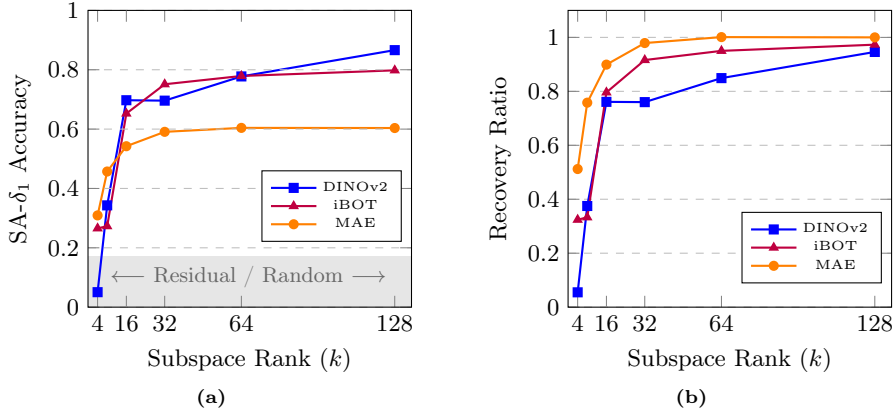

\begin{figure}[t] 
\centering
\includegraphics[width=\textwidth]{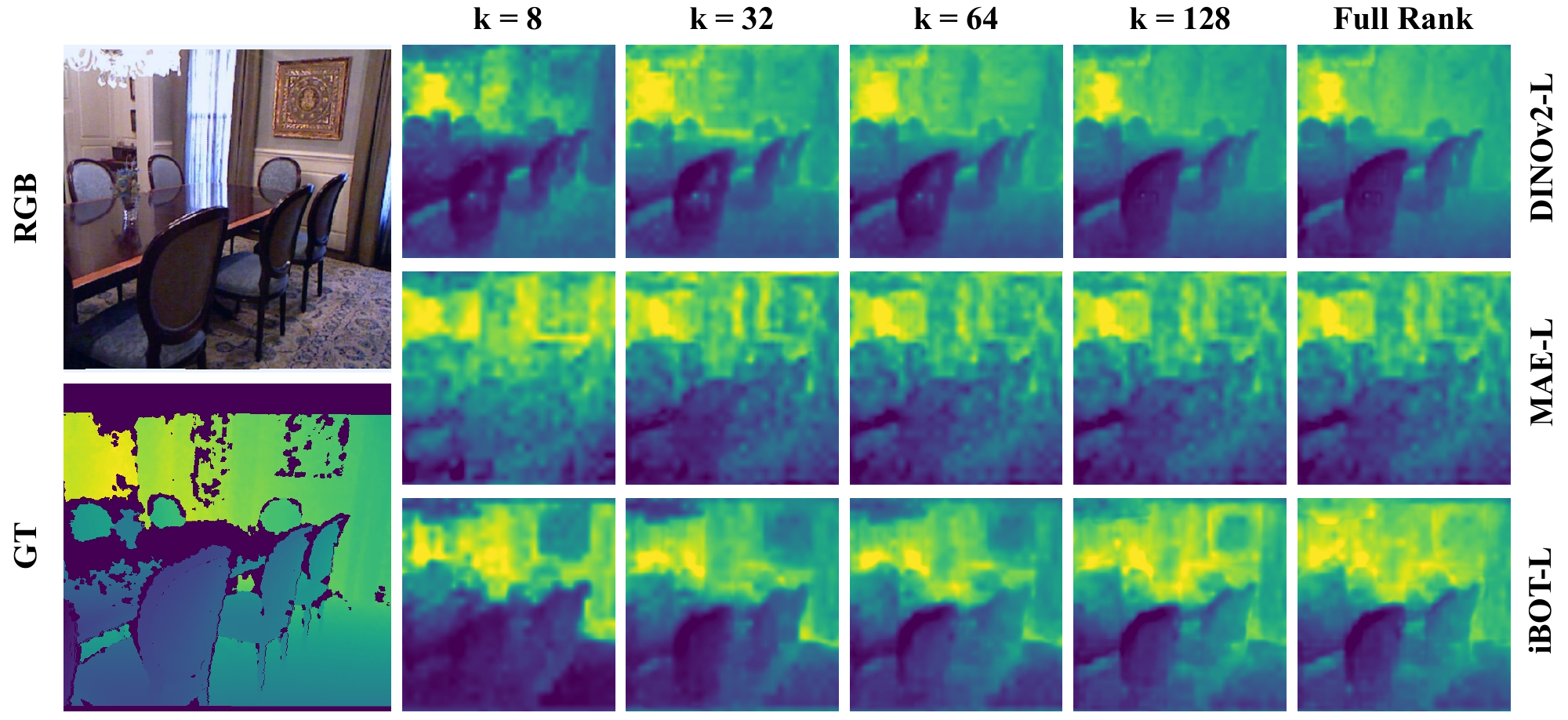} 
\caption{\textbf{Qualitative Visualization of Subspace Interventions.} We show depth predictions from the global linear probe across selected ranks ($k$). The patchy artifacts result from applying linear projections directly to coarse patch tokens, a deliberate choice intended to expose the raw feature structure. DINOv2 (top) recovers coherent scene layouts at an extremely low rank ($k=8$), demonstrating high linear compressibility. In contrast, MAE (middle) requires higher dimensional capacity ($k \ge 64$) to resolve basic object boundaries, as geometric signals are distributed among localized high-frequency details at low ranks. iBOT (bottom) represents an intermediate, hybrid state.}
\label{fig:vis_lowrank}
\end{figure}
\subsection{Global Compressibility and Energy Allocation}
\label{sec:global_subspace}
We apply subspace intervention on fused multi-layer representations to evaluate absolute saturation, relative structural efficiency, and energy allocation.

\subsubsection{High Compressibility of Geometric Representations.}
\label{sec:lowrank_core}
We truncate the task-aligned basis to isolate explicitly encoded geometric signals. \cref{fig:subspace_sig_noise} demonstrates that performance on both random ($\mathcal{R}_k$) and residual ($\mathcal{S}_k^\perp$) control subspaces collapses under the frozen linear probe. This gap indicates significant representational redundancy, suggesting that explicitly decodable geometric information can be projected into a low-rank subspace with minimal loss. This performance collapse confirms that the predictive mechanism of the original probe is strictly bottlenecked within the top-$k$ dimensions.

Qualitative visualizations (\cref{fig:vis_lowrank}) and recovery efficiency curves (\cref{fig:subspace_efficiency}) reveal distinct saturation behaviors across paradigms. MAE exhibits the fastest relative saturation, recovering over $98\%$ of its linear potential at $k=32$ despite its lower absolute geometric capacity. In contrast, DINOv2 requires a higher intrinsic dimensionality ($k \ge 64$) to converge. These results corroborate our accessibility findings. Specifically, MAE provides only coarse geometry in a linearly accessible form and requires few singular vectors for reconstruction, given that fine-grained details remain non-linearly entangled. In contrast, DINOv2 linearly aligns richer spatial details, which necessitates a broader basis to resolve precise object boundaries.

\subsubsection{Cross-Layer Energy Allocation.}
\label{sec:energy_allocation}

\begin{figure}[t]
\centering
\includegraphics[width=\textwidth]{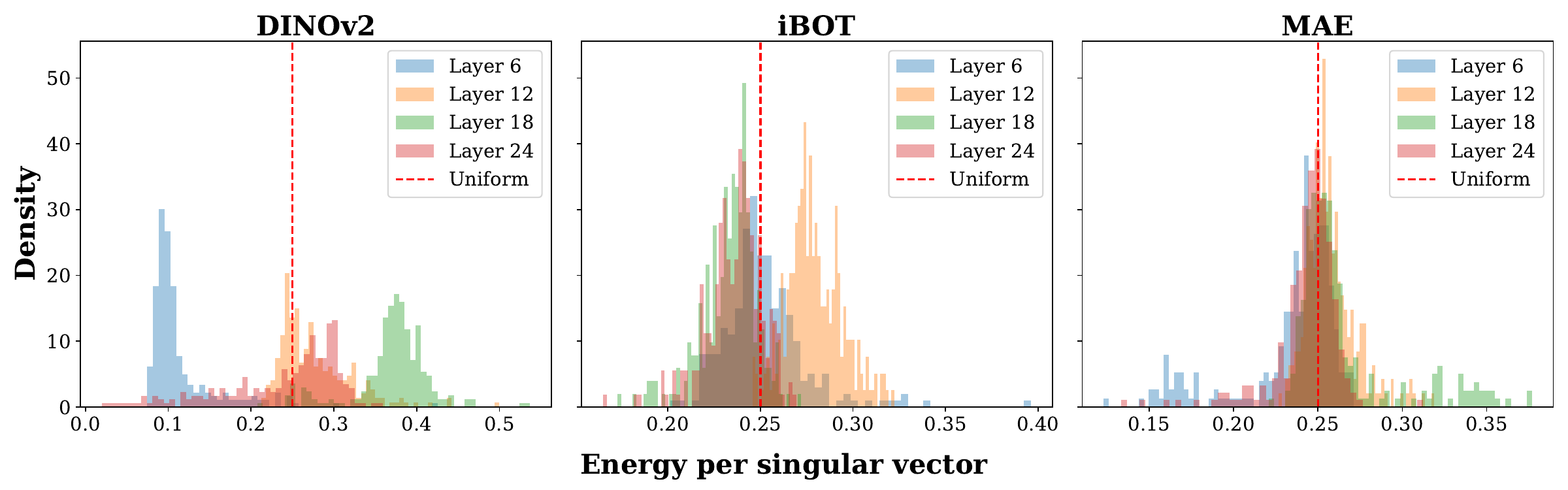}
\caption{\textbf{Energy Distribution Across Singular Directions.} Spectral-weighted energy density per singular vector. The dashed line represents the uniform baseline ($0.25$). DINOv2 shows multimodal peaks, indicating that geometric components are highly localized within intermediate layers. In contrast, iBOT and MAE exhibit overlapping distributions near the uniform baseline, indicating that their geometric signals are diffusely distributed across the hierarchy.}
\label{fig:energy_histogram}
\end{figure}

While global subspace analysis quantifies the capacity required for geometry, it does not identify which layers contribute most to this signal. To determine the contribution of each layer, we analyze the energy distribution of the converged global linear probe via its singular spectrum.

The probe maps concatenated features using weights $\mathbf{W}_{\mathrm{global}} \in \mathbb{R}^{C \times 4D}$. Each right singular vector $\mathbf{v}_m \in \mathbb{R}^{4D}$ resulting from the subspace decomposition encodes a task-aligned direction. We partition these vectors into four layer-specific blocks $\mathbf{v}_m = [\mathbf{v}_m^{(l_6)}, \mathbf{v}_m^{(l_{12})}, \mathbf{v}_m^{(l_{18})}, \mathbf{v}_m^{(l_{24})}]^\top$. To quantify the specific contribution of each depth, we define the spectral-weighted energy for layer $i$:
\begin{equation}
E_i = \frac{\sum_{m=1}^C \sigma_m^2 \|\mathbf{v}_m^{(l_i)}\|_2^2}{\sum_{j=1}^4 \sum_{m=1}^C \sigma_m^2 \|\mathbf{v}_m^{(l_j)}\|_2^2}
\end{equation}
where singular values $\sigma_m$ weight each direction by its explanatory variance.
\setlength{\intextsep}{0pt}
\begin{wraptable}{r}{0.45\linewidth}
\centering
\caption{Relative geometric contribution $E_i$ for four representative depths ($l_6$ to $l_{24}$).}
\label{tab:energy_allocation}
\setlength{\tabcolsep}{3pt} 
\footnotesize 
\begin{tabular}{lcccc}
\toprule
& \multicolumn{4}{c}{\textbf{Contribution (\%)}} \\ 
\cmidrule(lr){2-5}
\textbf{Model} & \textbf{$l_{6}$} & \textbf{$l_{12}$} & \textbf{$l_{18}$} & \textbf{$l_{24}$} \\ 
\midrule
\textbf{DINOv2} & 17.2 & 35.8 & \textbf{36.7} & 10.3 \\
\textbf{iBOT}   & \textbf{26.9} & 28.4 & 22.4 & 22.3 \\
\textbf{MAE}    & 19.5 & 28.4 & \textbf{32.7} & 19.4 \\
\bottomrule
\end{tabular}
\end{wraptable}

Spectral energy profiles (\cref{tab:energy_allocation,fig:energy_histogram}) reveal distinct routing strategies across depths. DINOv2 exhibits highly localized energy allocation, concentrating over 72\% of geometric energy within intermediate layers ($l_{12}, l_{18}$) before a sharp decline to approximately 10\% at $l_{24}$. This distribution suggests an empirical layer-wise affinity, where explicit geometric information is dominant in intermediate layers but declines in deeper ones. However, masked reconstruction and hybrid paradigms distribute this explicit signal. Both iBOT and MAE exhibit highly similar energy distributions that cluster near the uniform 0.25 baseline, requiring the linear probe to aggregate features across all depths simultaneously. This indicates that masked reconstruction distributes features across layers rather than concentrating them.

This global aggregation inherently blends distinct layer-wise states. For example, the multi-scale probe compensates for the final-layer drop in DINOv2 by relying on intermediate features, masking the abrupt loss of geometric capacity. To trace how spatial understanding evolves and identify where it degrades, we therefore transition to a strictly single-layer diagnosis.

\begin{figure*}[t]
\centering
\includegraphics[width=\textwidth]{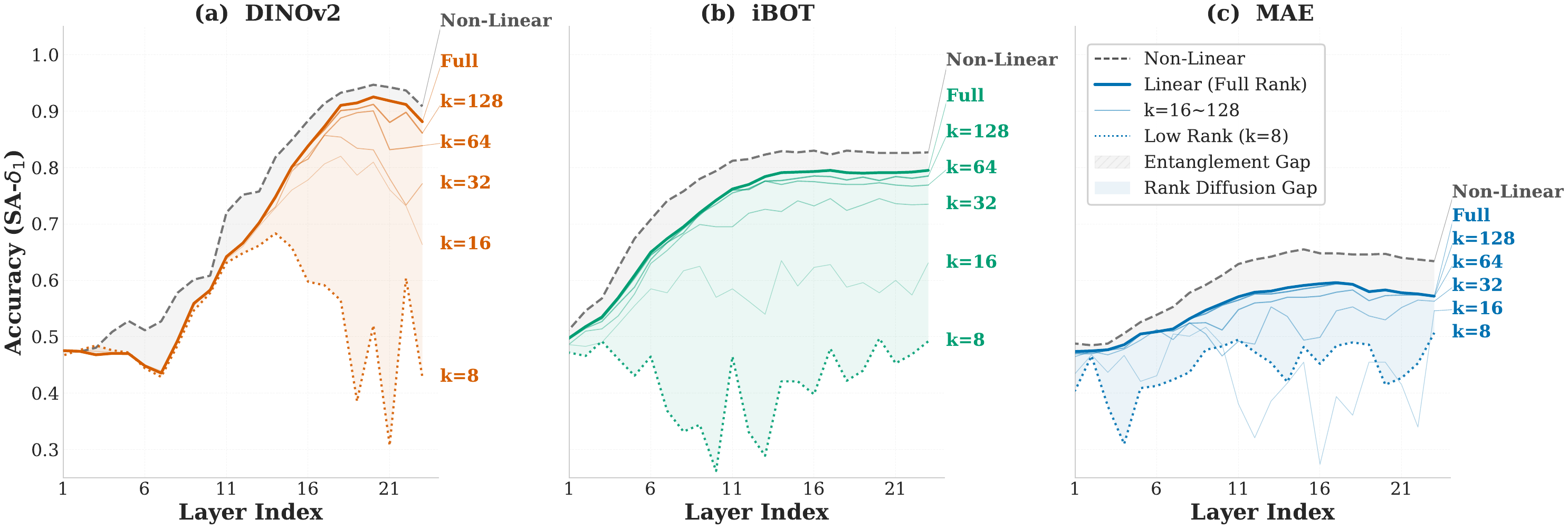}
\caption{\textbf{Layer-wise Subspace Analysis.} Performance evolution of geometric representations across different layers and subspace ranks. The Entanglement Gap highlights the disparity between linear and nonlinear decoding, indicating the degree of local non-linear folding. The rank dispersion gap highlights the performance drop under extreme low-rank constraints ($k=8$). Notice the distinct rank sensitivity transition in DINOv2 (a) at intermediate layers, contrasting with the continuous rank sensitivity in MAE (c) and iBOT (b).}
\label{fig:layer_rank}
\end{figure*}

\subsection{Layer-Wise Evolution and Rank Sensitivity}
\label{sec:phase_transition}

To trace the layer-wise evolution of geometric representations, we apply subspace interventions strictly to single-layer features. \cref{fig:layer_rank} tracks the performance trajectory from an extreme low-rank state ($k=8$) to the full linear capacity, benchmarked against a non-linear upper bound. We establish this non-linear baseline using a lightweight decoder incorporating spatial convolutions and residual connections on isolated single-layer features. Full architectural details are provided in the Supplementary Material.

As shown in \cref{fig:layer_rank}, DINOv2 exhibits significant geometric compactness in its early to middle layers. Up to layer 11, the $k=8$ subspace performs nearly as well as the full-rank linear probe, indicating that self-distillation effectively aligns spatial primitives into a highly compressed, linearly accessible format. A distinct transition occurs near layer 12, where the $k=8$ trajectory diverges from the full-rank baseline. This terminal decline in explicitly decodable geometry aligns with the localized intermediate energy routing observed in \cref{sec:energy_allocation}, implying a structural shift in the representational focus of the network. This transition motivates the subsequent analysis of layer-wise task affinities. In contrast, masked reconstruction distributes features across more dimensions. MAE shows immediate and persistent rank sensitivity, likely because pixel-level objectives preserve high-frequency textures that disperse geometric signals. Although iBOT exhibits some early-layer alignment, it does not achieve the low-rank compactness observed in DINOv2. Both masked reconstruction and hybrid models maintain a substantial entanglement gap across most depths, suggesting that their geometric information remains reliant on local non-linear structures.

Despite these divergent evolutionary paths, the $k=64$ and $k=128$ subspace trajectories closely track the full-rank linear baselines across nearly all depths in every paradigm. This consistency suggests a pervasive high compressibility, where explicitly decodable geometric representations can be recovered from a low-rank subspace regardless of the pre-training objective.

\subsection{Layer-wise Task Affinity}
\label{sec:semantic_tradeoff}

\begin{figure}[t]
\centering
\includegraphics[width=\textwidth]{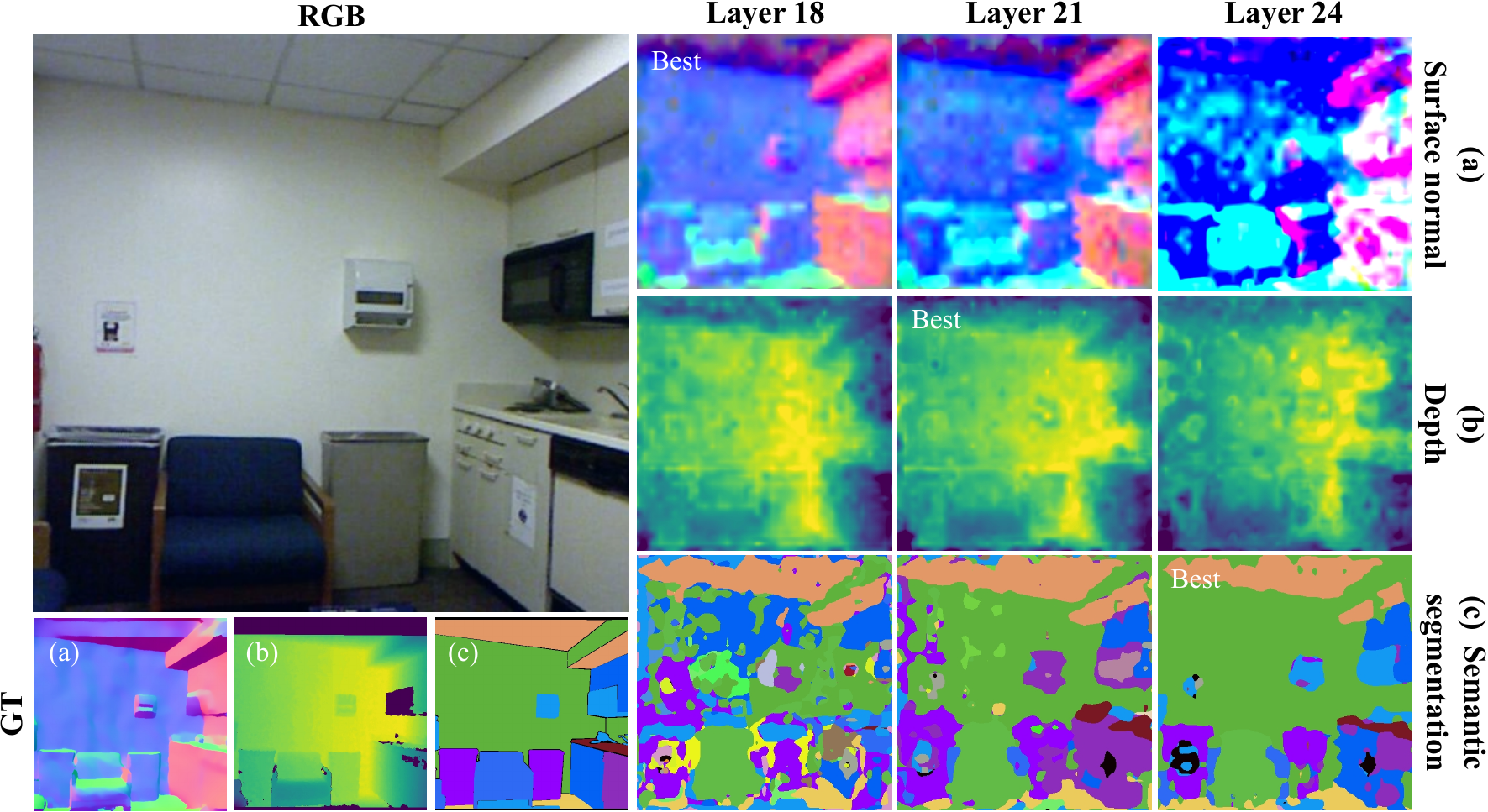}
\caption{\textbf{Qualitative Task Affinity in DINOv2.} Visual predictions decoded from frozen features at layers 18, 21, and 24. Performance for (a)surface normals, (b)depth, and (c)semantic segmentation peaks at these layers, respectively. This variation demonstrates distinct layer-wise affinities determined by the downstream task objectives.}
\label{fig:tradeoff_vis}
\end{figure}

Our layer-wise analysis reveals that the terminal layers of DINOv2 show a significant decline in explicitly decodable geometric information. To determine whether this reflects a loss of capacity or a shift toward semantic abstraction, we train identical linear probes for surface normal estimation, depth estimation, and semantic segmentation on NYUv2. Given that the input images and the frozen backbone remain constant, this controlled setup ensures that divergent layer-wise peak performances are driven by downstream task affinities. While this section primarily focuses on DINOv2, we extend this layer-wise analysis to MAE and iBOT. We observe a similar yet distinct trend: across all models, surface normal estimation consistently peaks in the earlier layers. However, the performance peaks for depth estimation and semantic segmentation do not exhibit a sharp demarcation. Detailed results and comprehensive comparisons are provided in the Supplementary Material.

\setlength{\intextsep}{0pt}
\begin{wrapfigure}{r}{0.48\textwidth}
  \centering
  \includegraphics[width=\linewidth]{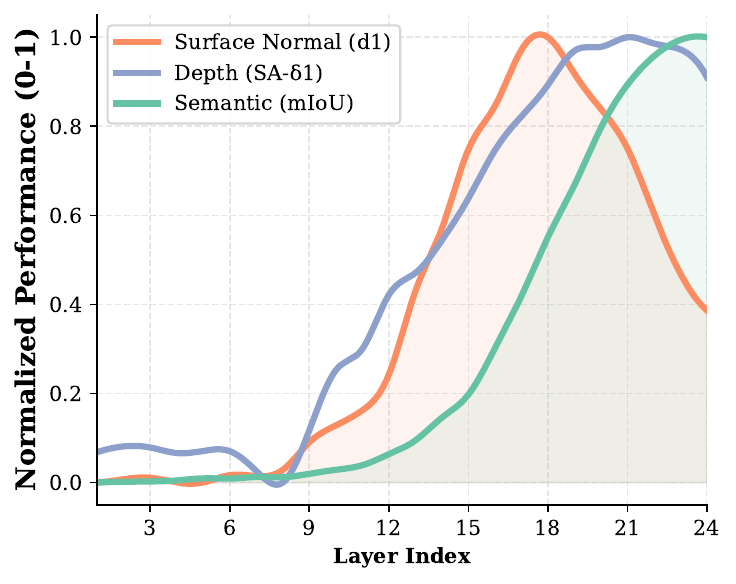}
  \caption{\textbf{Normalized probing performance.} Geometric tasks peak and decline early, whereas high-level semantic abstraction peaks latest.}
  \label{fig:tradeoff_line}
\end{wrapfigure}

As illustrated in \cref{fig:tradeoff_line}, the performance trajectories normalized against each task's layer-wise peak reveal a sequential transition between different tasks. Intermediate layers exhibit a strong affinity for local geometric features, with surface normal estimation peaking at layer 18, while depth estimation maintains structural accuracy longer and peaks at layer 21. This attenuation of explicit geometry is consistent with a steady improvement in semantic segmentation, which reaches its maximum at layer 24. Qualitative visualizations (\cref{fig:tradeoff_vis}) support this observation, showing that explicit geometric signals from intermediate layers are replaced by abstract representations in the terminal stages. This variation in task affinity suggests that relying solely on a terminal feature readout for multi-task dense prediction is sub-optimal, highlighting the need for depth-aware feature routing.

\subsection{Robustness of the Extracted Subspaces}
\label{sec:robustness}

\setlength{\intextsep}{0pt}
\begin{wrapfigure}{r}{0.48\textwidth}
  \centering
  \includegraphics[width=\linewidth]{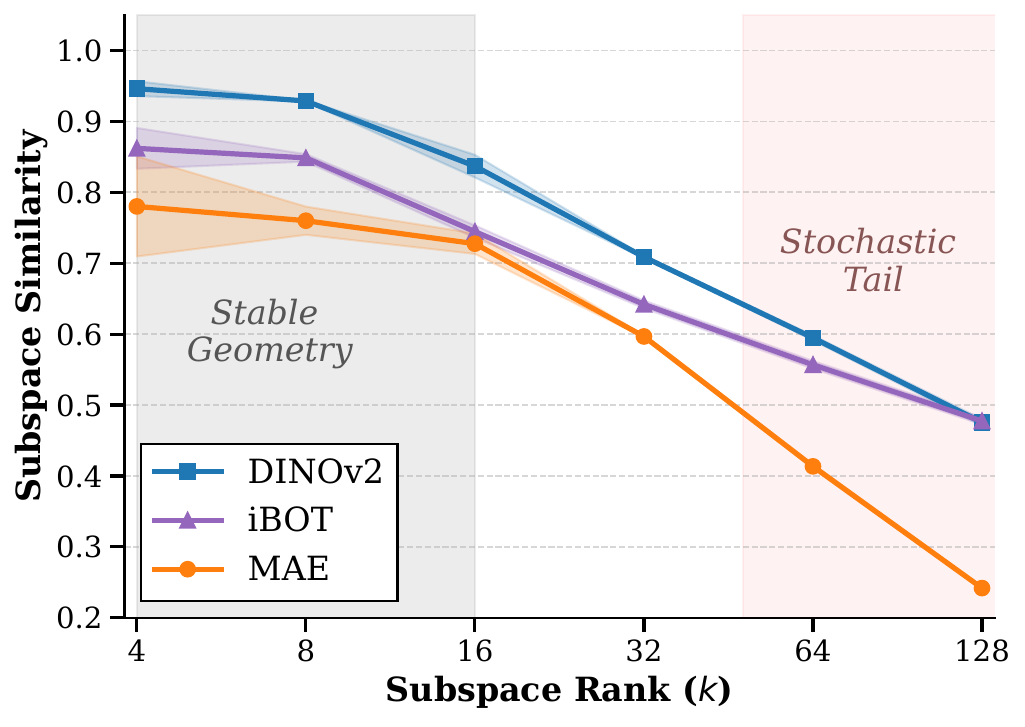}
  \caption{\textbf{Subspace Stability.} Lines and color shaded regions denote the mean and standard deviation of subspace similarity across three random seeds over ranks $k$. High consistency at low ranks ($k \le 16$) rules out optimization artifacts.}
  \label{fig:stability}
\end{wrapfigure}

To rule out optimization artifacts, we evaluate the subspace similarity of the extracted bases ($\mathbf{V}_k$) across three random probe initializations. \cref{fig:stability} reveals high consistency (\eg, $>0.93$ for DINOv2) at low ranks($k \le 16$), suggesting that the geometric core is a stable structural property rather than an optimization byproduct. Conversely, the low similarity in the tail dimensions ($k \ge 64$) indicates a lack of stable geometric signals, making these bases highly susceptible to random initialization. It aligns with the low-rank compressibility shown in \cref{fig:subspace_pair}. Detailed stability metrics are provided in the Supplementary Material.


\section{Related Work}

\subsection{Evolution of Probing Methodologies}
Self-supervised ViT models naturally learn robust semantic and geometric representations that transfer effectively to dense prediction tasks~\cite{Goyal2019ScalingAB, Caron2021EmergingPI, Oquab2023DINOv2LR, Ranftl2021VisionTF, Amir2021DeepVF}. While the broader advantages and limitations of these learning mechanisms have been systematically reviewed~\cite{Khan2024ASO}, standard probing efforts predominantly evaluate these models from a macroscopic perspective. For instance, recent studies confirm DINOv2's excellence in 2.5D view-centric geometry while revealing limitations in multi-view reasoning~\cite{ElBanani2024ProbingT3, Chen2024ProbingTM, Alam2026TheSB}. Crucially, these performance-driven evaluations treat the feature manifold as a black box, leaving the underlying organization of geometric primitives unexplained.

To address this limitation, recent research has shifted toward mechanistic interpretability. It is now established that visual concepts and spatial reasoning are not merely diffuse patterns but can be isolated into specific linear directions, attention heads~\cite{Bahador2025MechanisticIO}, or dissected via overcomplete dictionary learning~\cite{Fel2025IntoTR}. Our work extends this trajectory through controlled subspace intervention, directly isolating low-rank coordinate systems that encode explicit geometry.

\subsection{Subspace Analysis and Intrinsic Dimensionality}
The manifold hypothesis posits that high-dimensional visual data inherently resides on low-dimensional submanifolds~\cite{Pope2021TheID, Ansuini2019IntrinsicDO}. Empirical evidence confirms that highly parameterized pre-trained models exhibit remarkably low intrinsic dimensionality, yielding generalization bounds independent of their total parameter count~\cite{Aghajanyan2020IntrinsicDE, Li2018MeasuringTI}. This pervasive low-rank bias in deep networks causes increasing depth to drive representational compression, strictly bounding the number of non-negligible singular values~\cite{Patel2024LearningTC, Garrod2024ThePO}. In deep discriminative models, within-class variability of terminal-layer training activations collapses toward zero~\cite{Papyan2020PrevalenceON}. Specifically, Vision Transformers can experience exponential rank collapse within attention mechanisms, driving tokens toward a uniform state~\cite{Dong2021AttentionIN}.
 
Crucially, self-supervised learning (SSL) objectives govern the topology of these low-rank manifolds. Contrastive and self-distillation frameworks explicitly constrain representations onto a low-dimensional, hard-shell hyperspherical manifold~\cite{kumar2026learning}, effectively prioritizing invariant semantic concept extraction at the cost of spatial rigidity. In contrast, MAEs must preserve high-dimensional variance and high-frequency geometric primitives to satisfy dense pixel-level reconstruction constraints~\cite{He2021MaskedAA, Zhang2022HowMM}. Hybrid approaches combine these paradigms to balance global instance discriminability with fine-grained local awareness~\cite{Huang2022ContrastiveMA, Zhou2021iBOTIB}. Our analysis bridges these observations, revealing how distinct pre-training objectives dictate feature encoding formats and govern the layer-wise routing of explicit geometry and semantic abstraction.

\section{Conclusion}
This study explores the evolution of geometric representations in self-supervised vision models. Through targeted subspace intervention, we find that models like DINOv2 align geometric primitives into highly compressible, low-rank formats, whereas MAE disperses these signals across broader dimensions. Furthermore, the layer-wise analysis reveals a task affinity within deep representation spaces. We show that the decline of explicit geometric capacity in terminal layers is consistent with the emergence of peak semantic abstraction. By connecting these layer-wise representational shifts with downstream decoding complexity, our findings clarify the limitations of relying exclusively on terminal features for dense prediction. These findings provide a useful foundation for feature routing, more effective feature selection, and lightweight decoder design.

\subsubsection{Limitations and Future Work.} While our linear intervention rigorously isolates explicit geometric structures, it struggles to unroll highly non-linear feature entanglements. Furthermore, strict in-domain evaluation requires datasets with perfectly aligned depth, normals, and semantics. The scarcity of such comprehensive data limits broader outdoor or cross-domain validation. Additionally, due to resource constraints, we could not evaluate larger-scale models like ViT-G. Future work will explore non-linear topological interventions and synthetic multi-modal datasets to address these constraints.

\section*{Acknowledgements}
This work was supported by JST CREST, Japan, under Grant Number JPMJCR2554, and by JSPS KAKENHI under Grant Number JP26K02785.

%
\bibliographystyle{splncs04}
\bibliography{main}
\end{document}